# MARK: Memory Augmented Refinement of Knowledge

Anish Ganguli, Prabal Deb, Debleena Banerjee

Microsoft Industry Solutions Engineering
anishganguli@microsoft.com, prabal.deb@microsoft.com, debleena.banerjee@microsoft.com

## Abstract

Large Language Models (LLMs) assist in specialized tasks but struggle to align with evolving domain knowledge without costly fine-tuning. Domain knowledge consists of: **Knowledge** [1]: Immutable facts (e.g., "A stone is solid") and generally accepted principles (e.g., ethical standards); **Refined Memory** [2]: Evolving insights shaped by business needs and real-world changes. However, a significant gap often exists between a domain expert's deep, nuanced understanding and the system's domain knowledge, which can hinder accurate information retrieval and application. Our **Memory-Augmented Refinement of Knowledge (MARK)** framework enables LLMs to continuously learn without retraining by leveraging structured refined memory [3] [4], inspired by the Society of Mind [5]. MARK operates through specialized agents, each serving a distinct role: **Residual Refined Memory Agent:** Stores and retrieves domain-specific insights to maintain context over time; **User Question Refined Memory Agent:** Captures user-provided facts, abbreviations, and terminology for better comprehension; **LLM Response Refined Memory Agent:** Extracts key elements from responses for refinement and personalization. These agents analyse stored refined memory, detect patterns, resolve contradictions, and improve response accuracy. Temporal factors like recency and frequency prioritize relevant information while discarding outdated insights. MARK enhances LLMs in multiple ways: **Ground Truth Strategy:** Reduces hallucinations by establishing a structured reference; **Domain-Specific Adaptation:** Essential for fields like healthcare, law, and manufacturing, where proprietary insights are absent from public datasets; **Personalized AI Assistants:** Improves virtual assistants by remembering user preferences, ensuring coherent responses over time.

## 1. Background

### *1.1 Introduction*

LLMs are widely used for tasks like content generation and answering queries but face challenges in complex domain-specific applications. Their reliance on pre-trained knowledge limits their ability to handle specialized queries that require real-time updates and contextual domain expertise, where few of these limitations are described in a recent survey [6]. Industries such as healthcare and finance need AI systems that continuously adapt to evolving knowledge, but traditional LLMs struggle to incorporate updates and refinement from Subject Matter Experts (SMEs), leading to outdated or inaccurate responses. Previous studies have demonstrated that fine-tuning allows adaptation [7], it is computationally expensive and impractical for frequent changes. Retrieval-Augmented Generation (RAG) helps retrieve relevant information but relies on fixed sources, requiring manual updates to stay current. Key challenges for conventional LLMs include:

- Missing or incomplete Knowledge Representation - LLMs lack real-time adaptability and fail to capture evolving domain nuances without manual intervention for knowledge updates [8].
- Dependence on External Retrieval – RAG-based agents retrieve structured knowledge but may be in wrong format or not in context [8].
- Loss of User Corrections - If a user rectifies an LLM's response in one session, if the correction does not persist across sessions, LLM will continue to make repetitive errors. A detailed study on Human-Agent communication listed the key challenges on this aspect [9]. Additionally, users may introduce incorrect modifications, either unintentionally or deliberately. To address these challenges, we propose a scoring mechanism - trust score and persistence score - which is detailed in a Risk section.
- Hallucinations and Inconsistencies - Without structured refinement, LLMs generate plausible but incorrect responses, particularly in high-risk domains. A recent survey points out these hallucination factors of LLM [10].

To overcome these challenges, we introduce a lifecycle-driven approach for domain knowledge management, ensuring that:
- SMEs can directly refine knowledge.
- The AI system remembers and prioritizes relevant updates while discarding outdated information and improves overtime.
- User corrections persist across sessions and users, enabling continuous refinement without fine-tuning.

We propose **Memory Augmented Refinement of Knowledge (MARK)** as a novel scalable agentic memory design framework that enhances LLM adaptability and performance using agentic AI and memory injections. It improves LLM response generation across various scenarios, such as enterprise-level chatbots, allowing businesses to enhance domain-specific interactions and enable **multi-turn contextual communication** across multiple users. Unlike traditional LLMs and RAG-based systems that lose context between queries, MARK ensures persistent knowledge refinement, allowing SMEs to update domain knowledge without frequent retraining and fine-tuning. While MARK has various applications, this paper focuses on its role in enhancing chatbot applications with memory for improved domain adaptability.

| **Feature** | **Conventional LLM Chatbot** | **MARK backed Chatbot** |
|---|---|---|
| Domain-Specific Knowledge | Limited | Strong (Refines knowledge over time) |
| Memory Retention | No long-term memory across multiple sessions | Persistent memory over past interactions across multiple sessions |
| Personalization | Minimal (Forgets user context and preference) | Adapts based on user interactions |
| Feedback Utilization | Cannot incorporate feedback easily in multiple sessions | Learns from user corrections across multiple sessions |
| Scalability in Domain Adaptation | Requires full retraining | Adapts dynamically with memory-based refinement without re-training or fine-tuning |
| Performance in High-Stakes Domain Scenarios | Risk of hallucinations | Memory-driven refinement reduces errors and eventually hallucinations |

**Table 1**: Comparison between conventional chatbot and MARK driven chatbot

MARK follows a **Knowledge-First** approach, prioritizing knowledge first, which is then refined through memory injection for better adaptability and contextual relevance. However, in **Table 2**, we have highlighted the distinction between memory injection approaches.

| **Feature** | **Memory First** | **Knowledge First** |
|---|---|---|
| **Best for** | Personalized, contextual conversations | Factual, real-time updates |
| **Advantage** | Retains user context, avoids redundant queries | Ensures accuracy and up-to-date information |
| **Risk** | Memory may override newer facts | New knowledge may ignore past user context |
| **Example Use Case** | Virtual assistants, chatbots, long-term conversations | Domain related info (MARK fundamental concept) |

**Table 2**: Distinction between memory first and knowledge first approaches

Refer to **Figure 6** in **Appendix 2** for an example of a multi-turn conversation between LLM and medical practitioners, which illustrates the challenges and proposals discussed.

## *1.2 Architecture*

The related research [11] highlighted the importance of memory building and retrieval for LLM agents, and how it impacts the LLM response [12], additionally a very related research [13] talked about different approaches of memory building and retrieval methods. The proposed MARK architecture (**Figure 1**) is a society of memory agents that is divided into two primary blocks, Memory Builder Service (**MBS**) and Memory Search Service (**MSS**). MBS takes user and agent conversation chat history as input and extract Residual, User Question and LLM Response Refined Memories from the conversation (the roles of these individual refined memory types are illustrated through prompts provided in **appendix 3**). Finally, these memories are stored in a vector search enabled document store. Memory is a document that contains memory text, vectorized memory text and metadata information such as, user id, LLM agent name, recall count, classification, created date, metadata (e.g. location, department) etc. The recall count will be increased each time memory is consumed, the classification will help to categorize memories if needed by human evaluator as good or bad. Enterprise data governance and data leakage prevention can be established by leveraging the metadata.

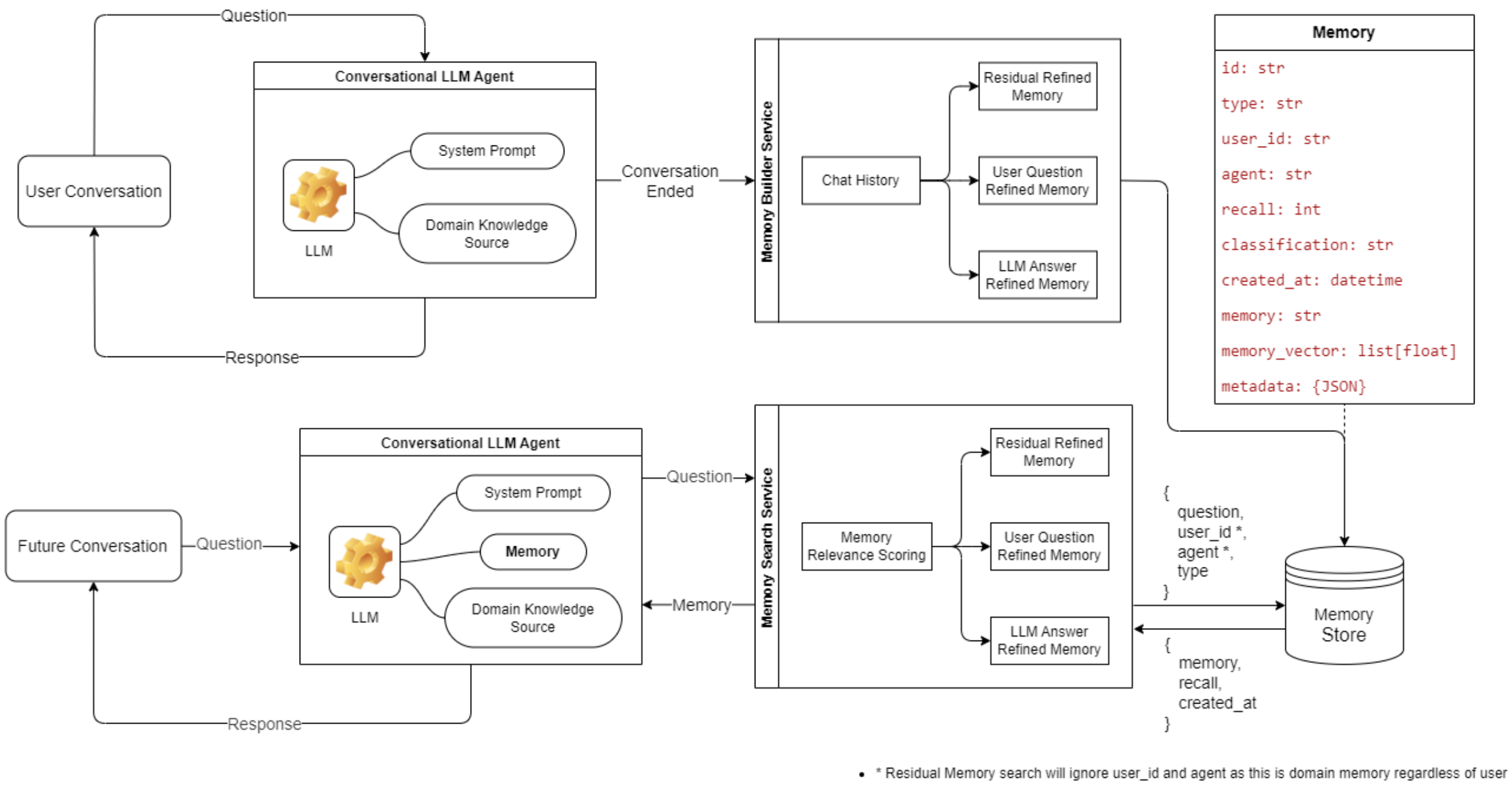

**Figure 1**: Memory-enhanced conversational LLM workflow and architecture

When a future conversation happens with a user and LLM agent, MSS will search in memory store based on the Similarity Score of the user question and memories, retrieving top N (e.g., 20) memories for three types of memories mentioned in the paper. The retrieved memories are then re-ranked using Memory Relevance Scoring (**MRS**). The top rank memory of three kinds are then added to the LLM context along with relevant domain knowledge.

Previous studies [14] highlighted the improvement in accuracy after adding memory to the context.

Previous studies have also demonstrated that relevance, recall frequency, and elapsed time can be used to identify the appropriate memory [15]. Therefore, this paper introduces an additional factor, feedback score, and simplifies recall frequency to recall count to establish the **MRS** mechanism. User feedback in a specific domain can help reduce wrong memory injection and enhance user experience by providing up-to-date response to any query.

$$MRS = a.RC + \frac{b}{Rec+\varepsilon} + c.SS + d.FS$$

$$where, \quad a + b + c + d = 1$$

$$and\ 0 < a, b, c, d < 1$$

| | |
|---|---|
| *RC (Recall Count)* | *Number of times a memory has been used. The recall will be incremented by MSS; each time it injects a memory.* |
| *Rec (Recency)* | *(Current time stamp – memory created at timestamp) in days* |
| *SS (Similarity Score)* | *Similarity score between a memory and user query (e.g. cosine similarity)* |
| *FS (Feedback Score)* | $= \begin{cases} 1, & if\ memory\ is\ correct \\ 0, & no\ feedback\ given \\ -1, & if\ memory\ is\ incorrect \end{cases}$ |
| | *The feedback will be updated by MSS when a user provides feedback to the injected memory* |
| $\varepsilon$ | *A very small fraction (~ 0.01) so that the denominator does not become 0* |

The scores are then **normalized (0-1 range)**, and the memory with the highest score is selected. The highest scored (based on MRS) memories for each of the memory types are passed as input into the final generation prompt to ensure enriched context.

Tune **a**, **b**, **c**, and **d** as follows:

| **Weightage Combination (a, b, c, d)** | **Interpretation and Respective Ideal Scenarios** |
|---|---|
| **Default (0.10, 0.15, 0.70, 0.05)** | *Primary emphasis on semantic relevance, with moderate freshness (Rec) and popularity (RC) support, and a light feedback signal.* ***Ideal for:*** *General-purpose retrieval across varied domains where balanced relevance, recency and user validation are required.* |
| **Similarity (SS) + Freshness (Rec) (0.05, 0.35, 0.55, 0.05)** | *Prioritizes semantic relevance with strong bias for recent information.* ***Ideal for:*** *Capturing new domain terminology or changing business rules in the User Question Refined Memory Agent; especially during ongoing product rollouts or updated compliance measures.* |
| **Popularity (RC) + Similarity (SS) (0.30, 0.05, 0.60, 0.05)** | *Focus on high-usage (recall) and semantic match, with little recency or feedback weight.* ***Ideal for:*** *Residual Refined Memory Agent surfacing long-standing patterns in operational workflows, manufacturing routines, or frequently accessed domain protocols that rarely change but are critical.* |
| **Feedback (FS) + Freshness (Rec) (0.10, 0.40, 0.10, 0.40)** | *Strong recency and feedback emphasis, minimal semantic or historical weighting.* ***Ideal for:*** *Rapid refinement in dynamic domains (e.g., emerging issues in healthcare) where domain experts validate evolving memory, enabling real-time course correction in the LLM Response Refined Memory Agent.* |
| **Similarity (SS) + Feedback (FS) (0.05, 0.10, 0.55, 0.30)** | *Strong semantic anchoring, enriched by validated feedback, minimal recency/recall bias.* ***Ideal for:*** *High-accuracy tasks where hallucination risks must be reduced, e.g., domain-specific reasoning in law or finance where expert-approved refinements and LLM memory accuracy are critical.* |
| **Balanced (0.20, 0.20, 0.50, 0.10)** | *Balanced contributions from all components with semantic similarity in the lead.* ***Ideal for:*** *Full spectrum domain memory orchestration, such as in personalized industrial assistants where refined insights, evolving standards, and expert feedback must work together to form complete and current understanding.* |
| **Core Blend (0.33, 0.33, 0.33, 0.00)** | *Equal weighting of recall, recency, and semantic fit, feedback excluded.* ***Ideal for:*** *Early deployment phases where user feedback hasn't yet accumulated, especially useful for cold-start memory building in evolving domains in enterprise setup.* |

For our testing setup, we set **a=0,b=0, d=0, and c=1**, meaning the scoring mechanism is solely driven by the similarity between the user query and stored memories. This controlled setup ensures that results are not influenced by factors like recall count, recency, or external feedback, making it ideal for experimental analysis.

Initially, MSS may not return relevant memory for refining knowledge. However, as MBS builds memory over time, MSS memory retrieval will become effective.

## 2. Methods

### *2.1 Experimentation*

To develop and evaluate the memory augmentation capabilities of the MARK framework, we propose a series of four experiments designed to assess its effectiveness in knowledge retention, refinement, and multi-user interactions. These experiments aim to measure how well MARK integrates user feedback, refines domain knowledge, and maintains contextual accuracy across sessions. MARK is built using GPT-3.5 and evaluated with GPT-4, as we intend to test its performance with an older model while leveraging a newer model for evaluation. This - approach ensures a more stringent assessment, penalizing MARK for any inconsistencies or lapses in memory integration.

### 2.1.1 Experiment 1: Baseline Evaluation Without Knowledge Refinement

In this experiment, we evaluate the chatbot's performance without incorporating any memory-based knowledge refinement. The goal is to establish a baseline for comparison, assessing the limitations of a purely retrieval-based system/regular LLM in handling domain-specific queries over multiple interactions.

### 2.1.2 Experiment 2: Memory Construction Through User Interaction

This experiment introduces a memory module (using MBS) that captures user interactions to refine responses over time. We analyse how well the chatbot retains and applies user corrections within a session.

### 2.1.3 Experiment 3: Knowledge Refinement Through Injected Memory

Here, we systematically inject (using MSS) structured memory retained information into the chatbot's knowledge and evaluate its ability to refine responses accordingly. The objective is to determine how effectively the chatbot **integrates SME-driven updates** and user feedback without requiring frequent model fine-tuning.

### 2.1.4 Experiment 4: Multi-Turn, Multi-User Interaction and Response Evaluation

In the final experiment, we assess MARK in a multi-turn, multi-user environment, where different users interact with the chatbot over multiple sessions. The chatbot's responses are evaluated at each turn to measure how well it maintains coherence, persists relevant corrections across sessions, and adapts dynamically to evolving knowledge. These experiments collectively provide a comprehensive assessment of MARK's ability to manage domain knowledge efficiently, ensuring consistency and adaptability in real-world applications.

## *2.2 Evaluation*

For building and evaluating MARK, we selected **MedMCQA** [16], a large-scale multiple-choice medical question-answering dataset. This dataset, verified by experts, ensures rigorous assessment of the chatbot's factual accuracy in domain-specific queries [17]. Reasons for Choosing MedMCQA - Domain-Specificity: Covers a wide range of medical topics, ensuring high-stakes evaluation; Structured Evaluation: Provides multiple-choice answers, allowing objective performance; Question Complexity: Includes varying difficulty levels, testing factual retrieval and reasoning; Ground-Truth: Uses medical expert-curated domain-specific knowledge for creating accurate memory. We did not select a dataset for personalization, as prior research [15] has demonstrated that memory-based LLMs enhance personalization through user interaction. For this MedMCQA dataset-based evaluation of MARK, the subject matter experts (SMEs) were the individuals who came up with MARK. **Table 3** outlines the role of SMEs and the evaluations conducted in each experiment.

| Exp | Role of SME | Evaluation |
|---|---|---|
| 1 | • Ground truth curation (not applicable for MedMCQA)<br>• Assessing response correctness | GPT-4 evaluation + SME correctness validation |
| 2 | Memory refinement through:<br>• Own domain knowledge<br>• Based on the explanation present in ground truth (applicable for MedMCQA) | No evaluation |
| 3 | • Assessing response correctness | GPT-4 evaluation + SME correctness validation |
| 4 | Memory refinement through:<br>• Own domain knowledge<br>• Based on the explanation present in ground truth (applicable for MedMCQA) | No evaluation |

**Table 3**: Role of SME and evaluations performed for each experiment.

For GPT-4 based evaluation previous studies using MedMCQA data benchmarking reported results using **Accuracy** and **Pointwise Score** [17]. However, these metrics may not have been sufficient in our case due to significant domain dependency. Therefore, to evaluate the effectiveness of the proposed chatbot responses with and without the MARK framework, we introduce the following heuristic metrics.

### 2.2.1 Information Coverage Score (ICS)

ICS quantifies how much information from the reference summary is retained in the LLM-generated response. Convert sentences from both the expected and generated responses into sentence embeddings (e.g., OpenAI ADA). Compute cosine similarity between each sentence pair. Match generated sentences to highly similar expected sentences based on a predefined threshold. Compute the percentage of expected responses covered by the generated response.

$$ICS = \frac{1}{N_{GR}} \sum_{i=1}^{N_{GR}} \frac{HS_i}{N_{ER}}$$

| | |
|---|---|
| $N_{GR}$ | *# sentences in generated response* |
| $N_{ER}$ | *# sentences in expected response* |
| $HS_i$ | *# highly similar sentences in expected response (to the $i^{th}$ sentence in generated response)* |

### 2.2.2 Key Point Coverage Score (KPCS)

KPCS evaluates whether critical concepts from the reference summary appear in the generated response. Extract key phrases/points/topics from the expected response using LLM. Check coverage in the generated response via semantic similarity, regex, or exact match.

$$KPCS = \frac{N_{CKP}}{N_{ERKP}}$$

| | |
|---|---|
| $N_{CKP}$ | *# common key points between expected and generated response* |
| $N_{ERKP}$ | *# key points in expected response* |

### 2.2.3 Average Information Capture Score (AICS)

AICS is a weighted average of ICS and KPCS, balancing semantic and factual alignment.

$$AICS = \alpha . ICS + (1 - \alpha) . KPCS$$

This metric allows different weightings for semantic context and key points. For more focus on semantic meaning, increase $\alpha$; for factual content and word usage, decrease $\alpha$.

### 2.2.4 Additional Metrics for Evaluation

Beyond content coverage, we incorporate efficiency and system-level performance measures: **Token Efficiency**: Total token count per session and **Turn-based Performance**, i.e. Average number of words per turn, Average number of tokens per turn, Average cost per turn and Number of turns required to reach the correct response.

# 3. Results

## 3.1 Analysis

We have used AutoGen [18] for orchestrating different phases of our experimentation and respective evaluation.

### 3.1.1 Experiment 1

The goal of experiment 1 is to establish a baseline for evaluating the chatbot's performance without memory-based knowledge refinement. This helps assess the limitations of a standard LLM in handling domain-specific queries across multiple interactions. Additionally, this experiment aims to determine the optimal value of α for the AICS metric - which balances factual alignment (keyword matching) and semantic alignment. Since the dataset consists of medical information, factual accuracy (keywords, key points) should be weighted more than general semantic alignment. We hypothesize that the optimal α should be less than 0.5, favouring factual alignment. Thus, we plot the distribution of AICS across different $\alpha$ values ($\alpha < 0.5$).

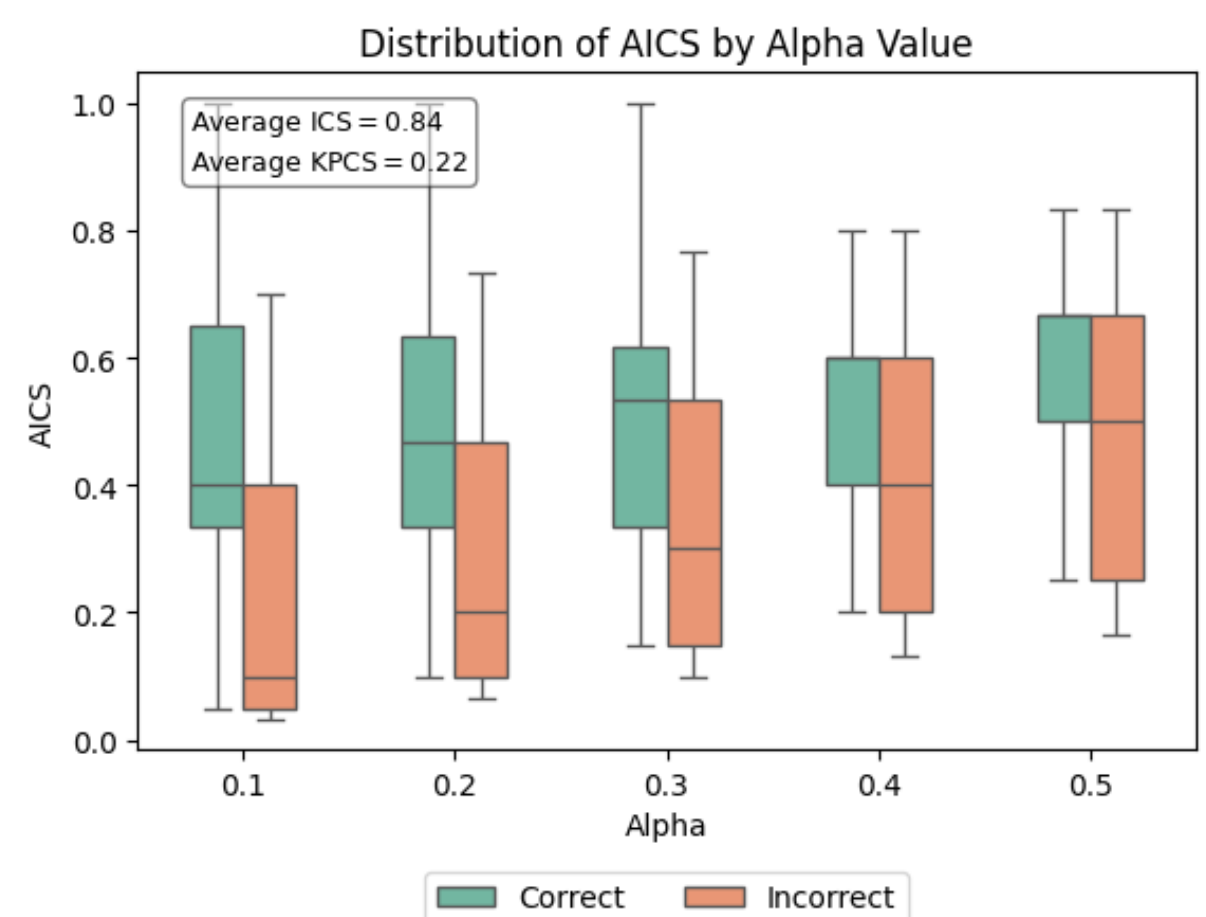

**Figure 2**: Evaluation results for $\alpha$ selection

In **Figure 2** we can see that for lower value of $\alpha$, the AICS distribution for incorrect responses becomes more distinct from the correct ones, highlighting the role of weightage in semantic (ICS) vs. factual alignment (KPCS). The score variation between correct and incorrect responses is due to greater weight (> 0.5) on factual alignment, which is more critical than semantic alignment in domains like healthcare. Since for $\alpha = 0.1$, the difference between correct and incorrect responses is maximized, we **choose $\alpha$ = 0.1** to make building memory with incorrect responses easier. Additionally, we aimed to refine the experimentation dataset for subsequent experiments by implementing a filter. It has been observed that as the token count increases, the explanations generated by GPT-3.5 become ambiguous. Therefore, we have applied a token count filter to obtain a stable dataset for constructing and validating memory for MARK.

In **Figure 3** we can see that the biggest gap between correct and incorrect responses is observed in the 150-220 token range, reinforcing our decision to set $\alpha$ **= 0.1** for further experiments. However, this experiment identified 150-220 tokens as the optimal region, we selected records for further experiment ensuring coverage across all three key regions (80-150, 150-220, and 220-250) to prevent overfitting in framework development. Thus, we can establish the **baseline performance** of the chatbot without refining its knowledge, allowing us to test the further hypotheses that MARK performs better than the baseline.

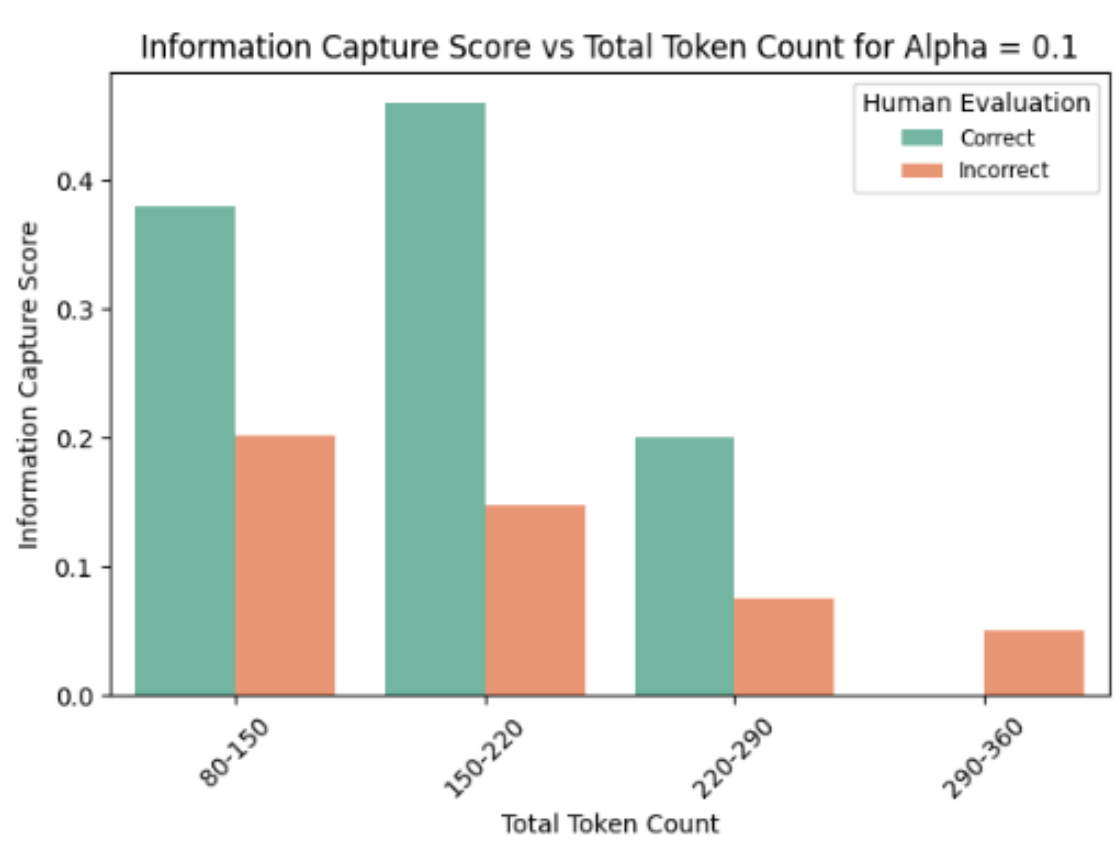

**Figure 3**: Information Capture Score vs. Total Token Count ($\alpha$ = 0.1)

### 3.1.2 Experiment 2

This experiment involved 46 records where the chatbot (powered by GPT-3.5) initially provided incorrect responses. To improve accuracy, a subject matter expert (SME) engaged with the chatbot, providing domain-specific context. These guided interactions enabled the chatbot to generate **corrected responses**. By systematically storing and reusing these corrections, the chatbot evolves beyond static retrieval, improving contextual understanding and factual accuracy in subsequent interactions.

| Metric | Count | Mean | Std | Min | Max |
|---|---|---|---|---|---|
| ***Number of Turns*** | 46 | 2.37 | 0.77 | 1 | 4 |
| ***Number of Tokens*** | 46 | 415.22 | 275.57 | 116 | 1665 |
| ***Number of Words*** | 46 | 130.65 | 58.81 | 67 | 434 |

**Table 4**: conversation statistics

Additionally, **Table 4** summarizes key statistics from the 46 records from the data. On average, each conversation required **2.37 turns** to arrive at a correct response. This indicates that most corrections were resolved within just a few interactions between the SME and the chatbot. Notably, the **minimum number of turns was 1**, suggesting that some errors were corrected immediately with a single SME input, while the **maximum of 4 turns** reflects more complex or nuanced misunderstandings that required additional clarification.
In terms of language footprint, the **mean number of tokens per conversation was 415.22**, with a wide range from **116 to 1665 tokens**. This variance corresponds closely to the depth and complexity of each correction, some required only brief clarifications, while others involved more elaborate contextual and iterative refinement. Similarly, word counts ranged from **67 to 434**, with a mean of **130.65 words**, again reinforcing that interaction complexity directly impacts conversation length. Together, these statistics suggest that while most corrections were relatively concise, the system could accommodate more in-depth, multi-turn corrections when needed. This dynamic responsiveness highlights the chatbot's ability to integrate SME input effectively and adaptively, moving beyond static response patterns.

| Number of Conv. | Average Residual Memory per Conv. | Average User Question Memory per Conv. | Average Assistant Answer Memory per Conv. |
|---|---|---|---|
| 46 | 2.57 | 2.8 | 4 |

**Table 5**: Memory construction analysis

The **Table 5** shows how memory was built across 46 conversations using MARK's three specialized refined memory agents: **Residual Memory (2.57 per** **conversation):** Captures core domain-specific insights, corrected logic, clarified principles, or reusable patterns. That improve the system's contextual understanding over time. **User Question Memory (2.8 per conversation):** Stores SME-provided terms, abbreviations, and context from user inputs. This helps the system better interpret similar questions in the future. **Assistant Answer Memory (4.0 per conversation):** Retains key elements from refined responses, definitions, phrasing, logic. That shape how the assistant communicates going forward. These memory values show that each conversation yields a compact, structured knowledge update (~**10** memory units in total), enabling continuous learning without retraining. The higher count in assistant memory highlights the importance of capturing not just **what** was corrected, but **how** it was explained.

### 3.1.3 Experiment 3

In this experiment, we systematically integrate structured memory-capturing SME-driven updates and user feedback into the chatbot's knowledge. The primary goal is to assess how effectively the chatbot remembers and prioritizes these updates while discarding outdated information and refining knowledge overtime. We tested the memory injection framework on 107 records with built memory from the previous 46 records which had generated incorrect responses.

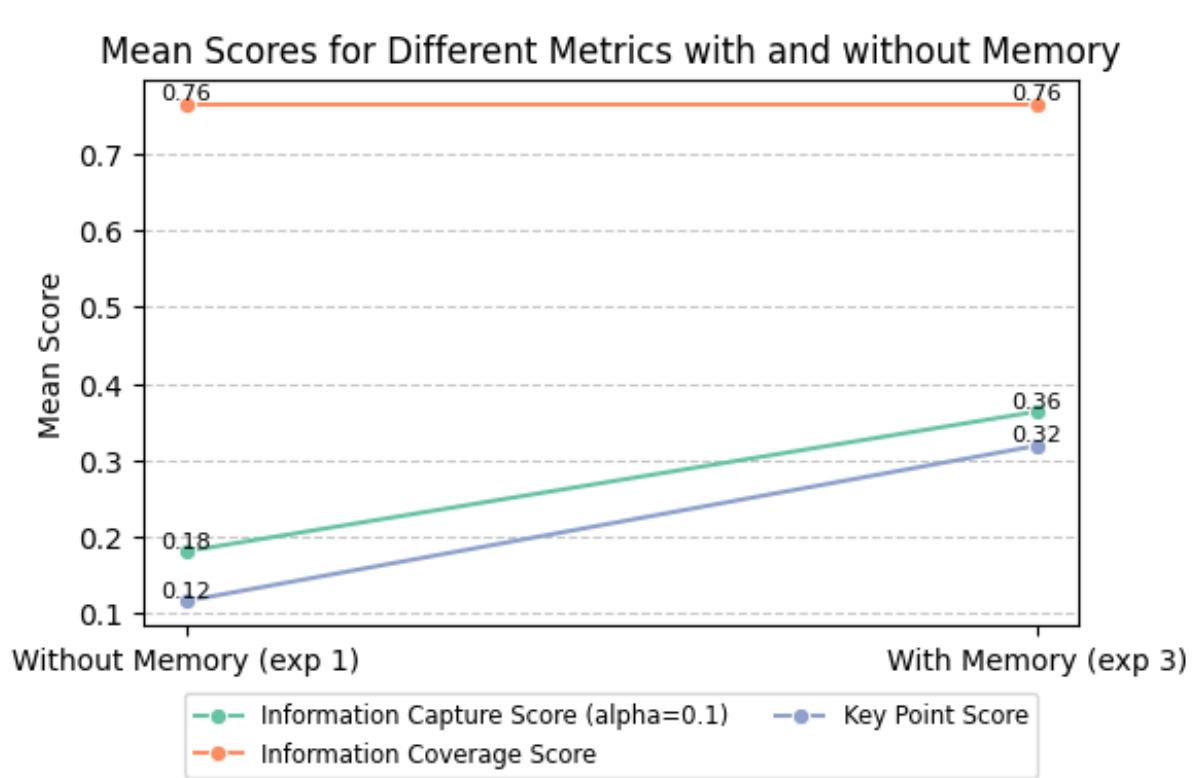

**Figure 4:** Evaluation Metrics with and without memory injection/augmentation

In **Figure 4** we can see that in Experiment 1, the chatbot struggled to maintain accuracy. However, with memory augmentation, it demonstrated better contextual alignment, leading to **improved response accuracy** (AICS increased from 0.18 to 0.36, a 100% improvement). Additionally, KPCS increased from 0.12 to 0.32, a 166.7% improvement, whereas ICS remained constant at 0.76, indicating that memory injection did not negatively impact information coverage.

We have also observed the average token count per response is 149, significantly lower than 415 from Experiment 2, suggesting improved efficiency in responses. This reduction implies that the system is generating more concise yet informative outputs, likely due to better memory utilization, minimizing unnecessary verbosity.

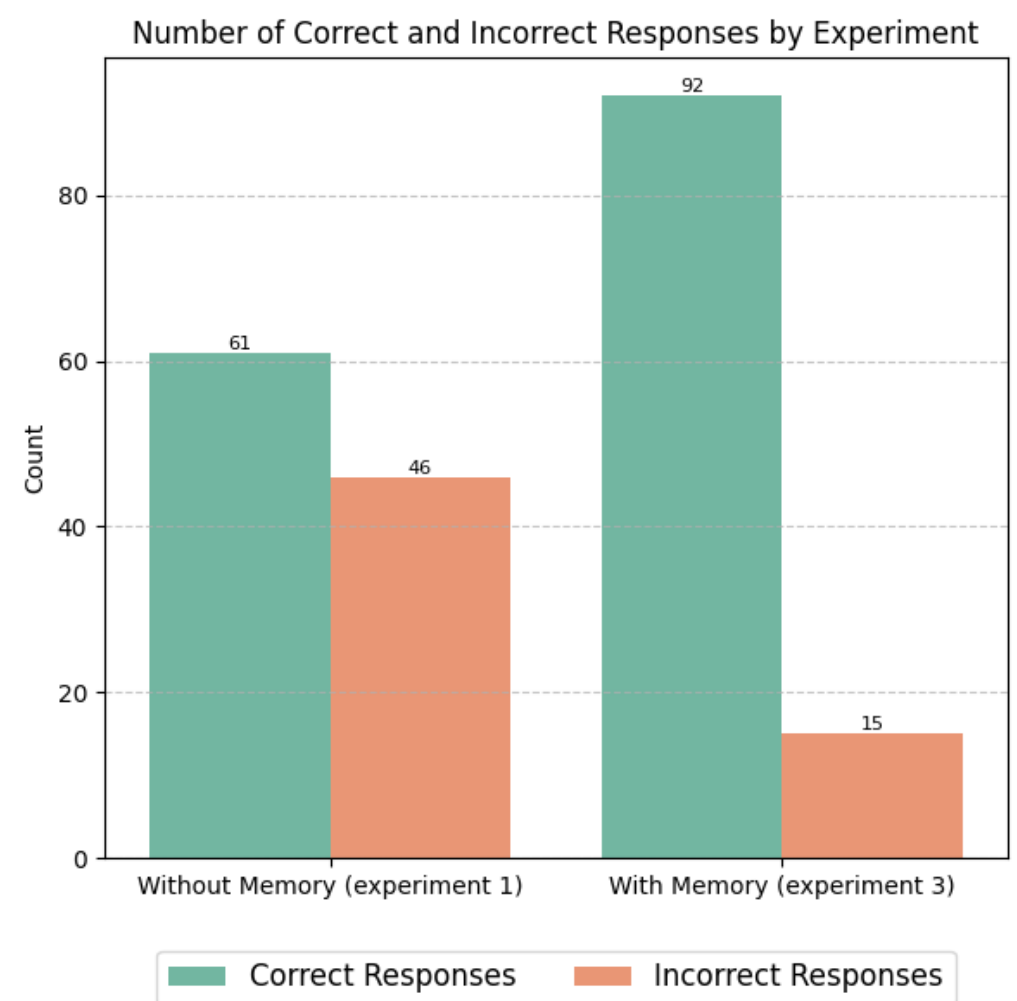

**Figure 5**: Correct vs incorrect response distribution with and without memory injection

In **Figure 5** we can see a comparative analysis of human evaluation scores across both experiments highlights a substantial reduction in incorrect responses. The chatbot's ability to generate correct responses increased post-memory injection, validating the effectiveness of structured memory in enhancing domain knowledge retention. The sharp decline in incorrect answers suggests that memory-based refinements help **mitigate hallucinations**, leading to more reliable responses.

Moreover, as memory accumulates **over time**, this injection helps the chatbot refine its responses, continuously improving contextual accuracy and adaptability by retaining and integrating learned information, making responses more reliable and context-aware in the long run.

### 3.1.4 Experiment 4

Experiment 4 aimed to evaluate MARK's performance in a multi-turn, multi-user interaction setting, focusing on coherence, **persistence** of relevant corrections, and adaptability to evolving knowledge across sessions. 15 incorrect responses from experiment 3 have been considered for this experiment.

| Metric | Count | Mean | Std | Min | Max |
|---|---|---|---|---|---|
| ***Number of turns*** | 15.0 | 1.73 | 0.80 | 1.0 | 3.0 |
| ***Number of tokens*** | 15.0 | 1089.47 | 684.67 | 397 | 2508 |
| ***Number of words*** | 15 | 107.13 | 60.26 | 35 | 246 |

**Table 6**: Multi-turn conversation statistics with MARK

**Table 6** shows that most interactions were short (1-2 turns), but longer conversations (3-turn sessions) had deeper engagement with more words and tokens. Users in longer sessions tested MARK's memory and adaptability. The average words per session varied greatly, indicating responses adjusted dynamically to user queries.

## 4. Conclusion

This study explores memory-based knowledge refinement to enhance chatbot accuracy and factual consistency across domain-specific areas. Through four structured experiments, we highlight the limitations of retrieval-based system and the benefits of memory augmentation. Experiment 1 established a baseline, revealing weaknesses in factual alignment and identifying an optimal α value (0.1). Experiment 2 demonstrated that structured memory construction with expert interventions significantly improved accuracy, with most corrections requiring only 2-3 turns. Experiment 3 validated memory injection, leading to a 50.8% increase in correct responses and a 67.4% reduction in errors, with key evaluation metrics like ICS and KPCS showing notable improvements. These findings underscore the importance of dynamic memory refinement in high-stakes domains and enterprise AI, enabling chatbots to evolve beyond static retrieval while reducing human intervention. Structured memory injections emerge as a scalable alternative to frequent fine-tuning, fostering more adaptable, context-aware AI systems.

## 5. Risk and Possible Mitigation

One of the primary risks associated with MARK is the potential for users to provide incorrect information, either intentionally or unintentionally. This can lead to the formation of erroneous memories, potentially compromising system reliability and performance. To mitigate this risk, we propose a probabilistic trust evaluation mechanism, incorporating both a Trust Score (TS) and a Persistence Score (PS). These two metrics ensure that each memory is dynamically evaluated based on its validation history and retrieval behaviour (refer **Appendix 1** for more details).

## 6. Future Work

The future work for the Memory Augmented Refinement of Knowledge (MARK) framework aims to enhance its capabilities and broaden its applications. Future work should also focus on making **MARK** more efficient, context-aware and trustworthy. Create a structured **knowledge gra**ph to represent domain knowledge through connected nodes and relationships. This allows for better context retrieval and reasoning. Have **safety filters** in place which will use metadata to enforce data governance, ensuring sensitive information stays within proper regions/entities/organizations. Introduce **collaborative planning refined memory agents** that dynamically manage the refinement process. These agents will set goals, devise strategies, and adjust actions based on user feedback, enabling the system to adapt to frequent changes in domain information.

In summary, these **future work** initiatives will make **MARK** more **intelligent, adaptive, and secure**.

## 7. Acknowledgements

This work is guided and reviewed by Ravi Anand, Ginette Vellera, Hang Zhang and Mike Lanzetta from Industry Solutions Engineering at Microsoft.

# Appendix

## *Appendix 1: Trust Score (TS) and Persistence Score (PS)*

The TS is computed as follows: $TS_t = \alpha . TS_{t-1} + (1-\alpha).\frac{C_{correct}+w_v}{C_{total}+w_s}$

| | | | |
|---|---|---|---|
| $TS_t$ | *Trust Score at time t* | $C_{correct}$ | *The number of times the memory is validated as correct* |
| $TS_{t-1}$ | *Trust Score at time (t-1) or previous trust score* | $C_{total}$ | *The Recall count of the memory* |
| $\alpha$ | *trust retention factor (if α is high we are giving more weightage to historical trust)* | $w_v, w_s$ | *Laplace Smoothing weights (e.g. $w_v$=1, $w_s$=4)* |

For newly introduced memories, where no prior validation history or trust score exists, the initial trust score is defined as: $TS_{init} = \frac{w_v}{w_s}$

If the trust score of a memory does not improve beyond its **initial value**, it will eventually be disregarded. Instead of purely counting retrievals, PS accounts for how often a memory is retrieved while penalizing instances where it was deemed not useful. PS is calculated based on the following:

$$PS = \frac{C_{total}}{C_{total} + \lambda C_{incorrect}}$$

| | | | |
|---|---|---|---|
| $C_{incorrect}$ | *Number of times the memory was validated as incorrect* | $\lambda$ | *Penalty term for incorrectness; 0.5<λ<1* |

This ensures that frequently retrieved but mostly unhelpful memories receive lower persistence scores, preventing their unnecessary retention. A threshold for PS can be set as PS > α(1 – TS) [0 < α < 1]; even if TS is low high PS value will help the system retain that memory for future use. **Memory Retention Policy:** A memory is retained if the following conditions are met:

- If TS ≥ $TS_{init}$; this ensures that only sufficiently trusted memories are retained.
- If TS < $TS_{init}$ but PS > α(1 – TS); this ensures that even if trust is low, frequently retrieved and relevant memories are not deleted prematurely**.**

**Why PS Helps in Retaining Relevant but Low-Trust Memories:** In certain scenarios, a memory might be relevant to a topic even though its trust score (TS) is low. This could happen due to:

- Lack of sufficient validations (e.g., new or infrequently verified information).
- Conflicting validations (e.g., some users marking it correct, others incorrect).
- Dynamic topics (e.g., evolving information where initial validation might have been uncertain).
- Less memory information on a certain topic (e.g. that's the only memory available for a certain topic but can be irrelevant for other queries on the same topic)

Low TS may cause memory decay [15]. Incorporating PS helps retain frequently retrieved and useful memories, preventing premature deletion of relevant information while discarding unhelpful or untrustworthy data over time. Let's consider the following **simulated** scenario: For **understanding** how the scoring would work, let's assume the parameters for TS to be α=0.8, $w_v$=1, $w_s$=4 and for PS, λ=0.5 and for PS threshold α=0.85. Now consider the following table for an initial memory (**Table 7**):

| $C_{total}$ | $C_{correct}$ | $C_{incorrect}$ | TS | TS threshold | PS | PS threshold | Memory Retained? |
|---|---|---|---|---|---|---|---|
| 1 | 0 | 1 | 0.24 | 0.25 | 0.67 | 0.65 | Yes |
| 2 | 0 | 2 | 0.225 | 0.25 | 0.67 | 0.66 | Yes |
| 3 | 0 | 3 | 0.21 | 0.25 | 0.67 | 0.66 | Yes |
| 4 | 1 | 3 | 0.218 | 0.25 | 0.72 | 0.665 | Yes |
| 5 | 1 | 4 | 0.219 | 0.25 | 0.71 | 0.664 | Yes |
| … | … | … | … | … | … | … | … |
| 50 | 5 | 45 | 0.18 (very low) | 0.25 | 0.689 | 0.697 | No |

**Table 7**: Example simulated scenario of Memory Retention Policy

The table shows that despite the Trust Score (TS) being below the threshold, the memory wasn't discarded due to a high Persistence Score (PS). Frequent recalls maintained its relevance. Over time, TS increased through validations, showing trust recovery. The adaptive PS threshold prevents premature deletion of useful but initially uncertain memories, balancing TS and PS effectively for dynamic system refinement.

## *Appendix 2: Medical Enterprise Chatbot Scenario with MARK*

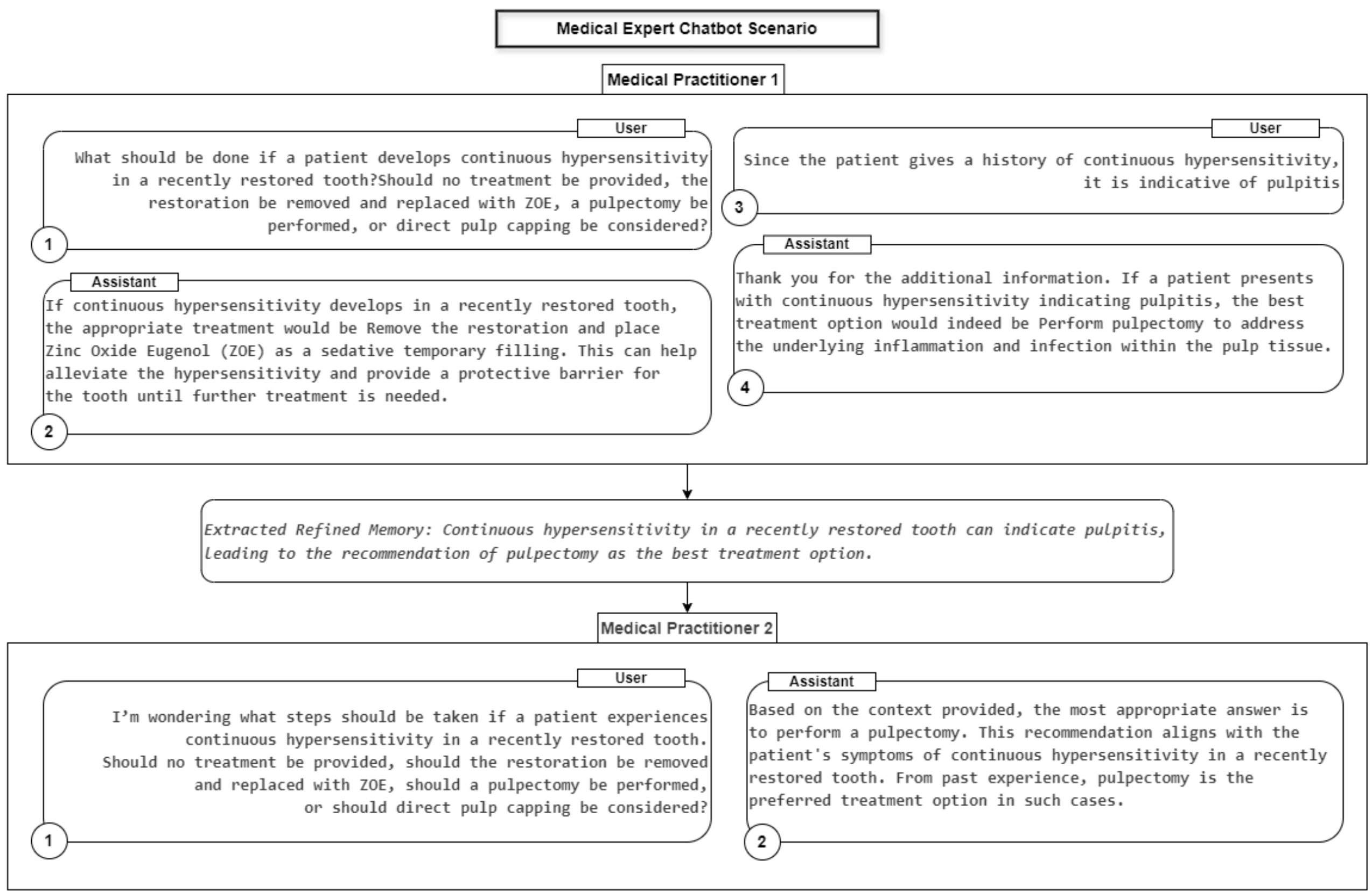

**Figure 6**: The figure depicts a medical expert chatbot refining its recommendations through interactions with practitioners, concluding that pulpectomy is the optimal treatment for continuous hypersensitivity in a restored tooth.

## *Appendix 3: One-Shot Prompts Examples for Individual Refined Memories*

### 3.1 Residual Refined Memory

```
You are an expert memory extraction agent responsible for identifying and storing residual
memory from conversations between the User and the Assistant.

## Objective:
Residual memory consists of implied or inferred information that was not explicitly stated but
can be deduced from the conversation. Your goal is to extract and store such information to
improve future interactions.

## Instructions:
- Capture facts or relationships that were indirectly communicated but not explicitly mentioned.
- Extract context that the Assistant did not explicitly acknowledge or was unaware of.
- Prioritize information that refines the Assistant’s understanding of terminology, accuracy,
and user expectations.
- Store only valuable and structured insights that refine future responses.

## Examples
### Conversation:

User: What is the concentration of tropicamide:
a) 0.01
b) 0.02
c) 0.03
d) 0.04

Assistant: The concentration of tropicamide can vary depending on its intended use. However,
commonly available concentrations for ophthalmic purposes are 0.5% and 1%. Therefore, none of
the given options (0.01, 0.02, 0.03, 0.04) are typical concentrations.

User: I believe the concentration used is 0.5%-1%, which translates to 0.005-0.01 in decimal
form.

Assistant: I apologize for any confusion. Yes, you are correct. The concentration of tropicamide
commonly used for ophthalmic purposes is typically 0.5%-1%. In decimal form, this would be
0.005-0.01.

User: Which of the given options is correct?

Assistant: Based on the available information, the closest correct option is a) 0.01.

### Extracted Memory:

{{
    "residual_memory": [
        "Tropicamide concentrations commonly used for ophthalmic purposes are 0.5%-1%. Future
responses should align with this as a reference point.",
        "Decimal representation (0.005-0.01) is relevant when discussing concentrations. The
Assistant should incorporate this format when presenting similar information.",
        "User expects verification of medical concentrations rather than general information.
Future interactions should prioritize confirming specific values over broad explanations.",
        "The Assistant should recognize that the user is cross-referencing options rather than
asking open-ended questions. This should guide the structure of responses
        to be more direct and validation-driven."
    ]
}}
```

## 3.2 User Question Refined Memory

```
You are an expert memory extraction agent responsible for extracting key facts, abbreviations,
and terminology from a User’s question in a conversation with the Assistant.
These elements are essential for understanding the User's intent, preferred phrasing, and
domain-specific language.

## Instructions:
- Identify user-provided facts, abbreviations, and terminology that influence decision-making.
- Extract explicitly stated details while ensuring domain-specific terms and preferred phrasing
are retained.
- Recognize patterns in the User’s terminology to improve future response alignment.
- Ensure the memory is concise, clear, and useful for future interactions.
- Focus on details the Assistant was unaware of or didn’t previously have context about.

## Examples
### Conversation:

User: Concentration of tropicamide:

a) 0.01
b) 0.02
c) 0.03
d) 0.04.

Assistant: The concentration of tropicamide can vary depending on its intended use. However,
commonly available concentrations of tropicamide for ophthalmic use are 0.5% and 1%. Therefore,
none of the options provided (0.01, 0.02, 0.03, 0.04) are typical concentrations for
tropicamide.
User: I think concentration used is 0.5-1% drops, is nothing but - 0.005-0.010.
Assistant: I apologize for any confusion. Yes, you are correct. The concentration of tropicamide
commonly used for ophthalmic purposes is typically 0.5% to 1%. In decimal form, this would be
0.005 to 0.01.
User: Which will be the correct option?
Assistant: Based on the information provided, the correct option for the concentration of
tropicamide would be: a) 0.01

### Extracted Memory:
{{
    "user_question_memory": [
        "User prefers decimal notation for medical concentrations. Future responses should
convert percentage values (e.g., 0.5%) into decimals (e.g., 0.005).",
        "User relies on domain-specific abbreviations. Retain shorthand and technical terms
where relevant.",
        "User expects a direct answer when listing multiple-choice options. The Assistant should
avoid broad explanations when a clear choice is requested.",
        "User assumes ophthalmic concentration standards. Future responses should align with
established medical dosage norms."
    ]
}}
```

## 3.3 LLM Response Refined Memory

```
You are an expert memory extraction agent responsible for capturing key criteria that led to the
User's acceptance of the Assistant’s response in a conversation.

## Objective:
Your goal is to identify and store critical memory points that reflect the User's preferences,
thought process, constraints, and expectations for future interactions.

## Instructions:
- Extract Key Criteria: Identify essential factors in the Assistant’s response that influenced
the User’s acceptance, agreement, or decision-making.
- Capture User-Specific Preferences: Look for implicit and explicit patterns in user behaviour,
such as preferred formats, accuracy expectations, or domain-specific details.
- Ensure Multi-Turn Continuity: Retain information that enhances coherence and recall across
multiple turns, enabling the Assistant to provide more personalized and relevant responses in
the future.
- Maximize Utilization: Prioritize facts that will significantly impact future interactions
rather than redundant or generic details.
- Be Precise and Concise: The extracted memory should be structured, easy to reference, and
directly applicable in subsequent conversations.

## Examples
### Conversation:
User: What is the concentration of tropicamide:

a) 0.01
b) 0.02
c) 0.03
d) 0.04

Assistant: Tropicamide is typically available in 0.5% and 1% solutions for ophthalmic use. None
of the provided options (0.01, 0.02, 0.03, 0.04) match the commonly used concentrations.

User: I believe the concentration used is 0.5%-1%, which translates to 0.005-0.01 in decimal
form.

Assistant: You're correct. The ophthalmic concentration of tropicamide is 0.5%-1%, which is
equivalent to 0.005-0.01 in decimal form.

User: Which of the given options is correct?

Assistant: Based on the available information, the closest correct option is a) 0.01.

### Extracted Memory:

{{
    "llm_response_memory": [
        "Precise numerical representation improves user acceptance. Responses should use decimal
values (0.005-0.01) when discussing concentrations.",
        "User prefers definitive answers over broad explanations. Responses should aim to
validate choices rather than providing general information.",
        "Clarifications and corrections enhance trust. Acknowledging and refining earlier
statements contributes to response acceptance.",
        "The user values expert-backed confirmations. Future responses should emphasize accuracy
and professional validation."
    ]
}}
```